\pdfoutput=1

\documentclass[11pt]{article}

\usepackage[final]{emnlp2021}

\usepackage{times}
\usepackage{latexsym}

\usepackage[T1]{fontenc}

\usepackage[utf8]{inputenc}

\usepackage{microtype}

\usepackage[linesnumbered,ruled]{algorithm2e}
\usepackage{amsmath}
\usepackage{algorithmic}
\usepackage{chngcntr}
\usepackage{graphicx}
\usepackage{float}
\usepackage{caption}
\usepackage{subcaption}
\usepackage{scrextend}
\usepackage{hyperref}
\usepackage{xcolor}

\title{{ARMAN}: {P}re-training with {S}emantically {S}electing and {R}eordering of {S}entences for {P}ersian {A}bstractive {S}ummarization}

\author{
  Alireza Salemi\textsuperscript{1}, Emad Kebriaei\textsuperscript{1}, Ghazal Neisi Minaei\textsuperscript{1}, Azadeh Shakery\textsuperscript{1,2} \\
  \textsuperscript{1}School of Electrical and Computer Engineering \\ College of Engineering, University of Tehran, Tehran, Iran \\
  \textsuperscript{2}School of Computer Science \\
  Institute for Research in Fundamental Sciences (IPM), Iran \\
  \small{\texttt{\{alireza.salemi,emad.kebriaei,ghazal.minaei,shakery\}@ut.ac.ir}}
}

\begin{document}
\maketitle

\begin{abstract}
Abstractive text summarization is one of the areas influenced by the emergence of pre-trained language models. Current pre-training works in abstractive summarization give more points to the summaries with more words in common with the main text and pay less attention to the semantic similarity between generated sentences and the original document. We propose ARMAN, a Transformer-based encoder-decoder model pre-trained with three novel objectives to address this issue. In ARMAN, salient sentences from a document are selected according to a modified semantic score to be masked and form a pseudo summary. To summarize more accurately and similar to human writing patterns, we applied modified sentence reordering. We evaluated our proposed models on six downstream Persian summarization tasks. Experimental results show that our proposed model achieves state-of-the-art performance on all six summarization tasks measured by ROUGE and BERTScore. Our models also outperform prior works in textual entailment, question paraphrasing, and multiple choice question answering. Finally, we established a human evaluation and show that using the semantic score significantly improves summarization results.
\end{abstract}

\section{Introduction}
Abstractive text summarization is the task of generating a short, fluent, and concise text that contains novel words and phrases other than the original document, preserving the primary subjects in the document. In contrast with extractive summarization, which aims to select the most important parts of the text to generate a summary, in abstractive summarization, the main goal is to generate a new persuasive piece of text as the summary of a document.


Earlier abstractive summarization works (\citealp{Herman}; \citealp{see-etal-2017-get}; \citealp{rush-etal-2015-neural}) focused on training with large datasets containing pairs of documents and summaries in a supervised manner. By introducing Transformer \citep{vaswani_2017} architecture and pre-training objectives and their positive impact on most NLP tasks, most current state-of-the-art (SOTA) methods focused on self-supervised objectives for pre-training Transformer architecture in abstractive summarization tasks \citep{liu-lapata-2019-text, PEGASUS, qi-etal-2020-prophetnet}. However, current pre-training  
works give more points to the summary with more words in common with the main text and pay less attention to
the semantic similarity between generated sentences and the original document. 

According to \citet{simons2017ethnologue}, the Persian language is one of the top 25 spoken languages in the world. However, there are limited research studies in Persian document summarization, and most of the prior works were mainly focused on extractive summarization. The main focus of this work is on Persian abstractive summarization. Nevertheless, our proposed method is language-independent. 

In this work, we first bring semantic similarity scores into a sentence selection schema to create a document's pseudo summary. Briefly, we prepare a summary corresponding to each document in a dataset by selecting important sentences based on semantic scores in a self-supervised manner. 
Next, we propose three novel objectives for pre-training a seq2seq Transformer. Our model, ARMAN, uses Transformer encoder-decoder structure and introduces a new combination of masking sentences with sentence shuffling and reordering objectives. We fine-tuned the models on six downstream tasks. According to an experiment, we found that letting the training model to copy pieces of the input text into the output summary does not lead to better results in downstream tasks. Experiment results showed that our proposed models obtained SOTA performance in all Persian abstractive summarization datasets on both ROUGE \citep{ROUGE} and BERTScore\citep{BERTscore}. Our models generated even better summaries than previous SOTA in zero and few shot settings when fine-tuned with a small number of document-summary pairs. We achieved SOTA results on two datasets with only 1K examples. 
Moreover, our proposed models performed well in other NLU tasks, including textual entailment, question paraphrasing, and multiple choice question answering. Finally, to ensure the significant improvement in summarization, we held a human evaluation, and we performed a student t-test on its results.

The main contributions of this paper are three-fold:
\begin{itemize}
    \item We introduce a top-sentence selection algorithm based on a semantic score to make document-summary pairs in a self-supervised manner.
    \item We propose three novel objectives to pre-train a Transformer encoder-decoder architecture for Persian abstractive text summarization that outperforms previous state-of-the-art models on six downstream tasks.
    \item We created an abstractive summarization dataset called Tebyan.
\end{itemize}

\section{Related Work}
Automatic text summarization was mainly performed based on statistical methods \citep{10.5555/1619499.1619564}; most of them were striving to rank sentences by extracting their features(\citealp{svore-etal-2007-enhancing}; \citealp{Erkan_2004}; \citealp{filippova-altun-2013-overcoming}). By rising of sequence to sequence learning with neural networks (\citealp{10.1162/neco.1997.9.8.1735}; \citealp{10.5555/2969033.2969173}) and attention mechanism \citep{Bahdanau2015NeuralMT} usage in abstractive summarization tasks \citep{nallapati-etal-2016-abstractive}, a new era in abstractive summarization began.

By introducing Transformer \citep{vaswani_2017} and Masked Language Modeling (MLM) methods of BERT \citep{devlin-etal-2019-bert}, most NLP tasks achieved a vast improvement gain using these pre-training methods and architectures. Following BERT's approach, many other Language Models were trained (\citealp{Liu2019RoBERTaAR}; \citealp{joshi-etal-2020-spanbert}) with differences in the amount of data used for pre-training and some optimizations on BERT's pre-training method; most of them were only Encoders.  Furthermore, Encoder-Decoder models were trained with a mixture of pre-training tasks; T5 \citep{JMLR:v21:20-074} and BART \citep{lewis-etal-2020-bart} are two of them.

Since the pre-training of Transformer was successful on most NLP tasks, some models were pre-trained for specific duties; PEGASUS \citep{PEGASUS} is a pre-trained model that was trained specifically for summarization on C4 and HugeNews corpora. PEGASUS trained with Gap Sentence Generation (GSG) that masks the most important sentences based on syntactic similarity of sentences of a document. ARMAN is different from PEGASUS in that we mask the most important sentences based on the semantic similarity of sentences. Furthermore, we use only a single mask token for any consecutive sentences that should be masked. This approach helps the model learn how many sentences should be generated for each masked token in the input sequence.
STEP \citep{zou-etal-2020-pre} is another pre-trained summarization model trained with MLM, Next Sentence Generation (NSG), and Sentence Reordering (SR) objectives. ARMAN uses SR as one of the pre-training methods in a modified form; we change the order of sentences in the input document. The model should select the most important sentences using semantic similarity of sentences to the document, then reorder them in the actual order that they appeared in the original document.

In the Persian language, some extractive summarization methods exist (\citealp{khademi2018conceptual}; \citealp{Rezaei2019FeaturesIE}; \citealp{8765279}; \citealp{Khademi2020}), but to the best of our knowledge, we know just one model on abstractive summarization. \citet{farahani2020leveraging} have used ParsBERT \citep{farahani2020parsbert} checkpoint with \citet{10.1162/tacl_a_00313}'s method to train a new sequence to sequence model with pre-trained weights for the encoder and decoder. In this regard, ARMAN is one of the first works on abstractive summarization for the Persian language. Also, ARMAN was able to achieve SOTA results on all available datasets.

\section{Methodology}
This section introduces a sentence selection method based on semantic similarity scores to make a pseudo summary. Then, we propose three novel objectives for pre-training a seq2seq model for the abstractive summarization tasks.

\subsection{Top Sentence Selection (TSS)}
We introduce a new semantic-based approach for selecting important document sentences to make a pseudo summary in this work. The pseudo summary consists of important sentences of a given document, and the models are supposed to generate an output similar to the pseudo summary corresponding to the document. For comparison, we also use a syntactic-based metric to select sentences from the original document. Inspired by recent work in generating pseudo summaries \citep{PEGASUS}, we select sentences from a document based on two strategies and concatenate them to create a pseudo summary. For each document in a data collection, we make a summary as described in Algorithm \ref{algo:seq}. At first, we calculate a score function for each pair of $(sentence, document\setminus sentence)$. Then we calculate the top $m$ sentences and merge them to make the pseudo summary. The parameter $m$ is calculated based on the number of sentences.
\begin{algorithm} 
\caption{Top Sentence Selection}
\label{algo:seq}
\SetKwInOut{Input}{Input}
\SetKwInOut{Output}{Output}
\Input{Document}
\Output{Text, Summary}
\begin{algorithmic}
\FOR{$s_i$ in $Document$}  
    \STATE $r_i := score\_func(s_i,Document\setminus s_i)$
\ENDFOR
\STATE $Summary := \emptyset$
\STATE $Text := Document$
\FOR{$j\leftarrow 1$ to $m$}
    \STATE $k := argmax\{r_i\}_{\forall\,s_i\,\notin Summary}$
    \STATE $Summary := Summary \cup \{s_k\}$
    \STATE $Text := Text \setminus \{s_k\}$
\ENDFOR
\end{algorithmic}
\end{algorithm}

\noindent\textbf{Syntactic-based approach}:
In this strategy, we create a pseudo summary by selecting and merging sentences from a document using a syntactic-based approach. ROUGE is a mainly used metric that calculates the similarity between a candidate sentence and a collection of reference sentences based on the overlap of N-grams \citep{ROUGE}. The higher the ROUGE score between two pieces of text, the more similar they are. The \textit{score-func} in Algorithm \ref{algo:seq} calculates the ROUGE1-F1 score between the sentence and remaining sentences of the document. PEGASUS \citep{PEGASUS} has used such a method as Gap Sentence Generation.

\noindent\textbf{Semantic-based approach}: Although selecting sentences based on the ROUGE metric is simple, cost-effective, and usable in low-resource languages, ROUGE comes with some drawbacks \citep{kryscinski-etal-2019-neural}. In particular, ROUGE does not account for different words with the same meaning since it only calculates syntactical matches. Thus, if we have two sentences with the same meaning but expressed with different words, they will be assigned a low ROUGE score.
To the best of our knowledge, this paper is the ﬁrst to study semantic similarity in creating pseudo summaries and its effect on the quality of generated summaries. 

To consider the semantic score in calculating the similarity of two sentences, we used a recent BERTScore metric. BERTScore computes a similarity score for each token in the candidate sentence with each in the reference sentence using contextual embeddings \citep{BERTscore}. Due to the high computational cost of calculating this metric for each pair of $(sentence, document\setminus sentence)$, we used FastText \citep{FastText} pre-trained embeddings instead of BERT contextual embeddings. According to BERTScore, for a reference $x$, and a candidate $\hat{x}$, the recall, precision, and F1 scores are:
\begin{gather*} 
    R_{FT} =\frac{1}{|x|} \sum_{x_i \in x}   \max_{\hat{x}_j \in \hat{x}} \mathrm{\textbf{x}_i}^\top \hat{\mathrm{\textbf{x}}}_{j},\\
    P_{FT} =\frac{1}{|\hat{x}|} \sum_{\hat{x}_j \in \hat{x}}   \max_{x_i \in x} \mathrm{\textbf{x}_i}^\top \hat{\mathrm{\textbf{x}}}_{j},\\
    F1_{FT} =2\frac{P_{FT}.R_{FT}}{P_{FT}+R_{FT}} .
\end{gather*}
For applying semantic score, the score function in Algorithm \ref{algo:seq} calculates $F1_{FT}$\footnote{FT stands for FastText.}. 

\begin{figure*}[!ht]
    \centering
    \includegraphics[scale=0.25]{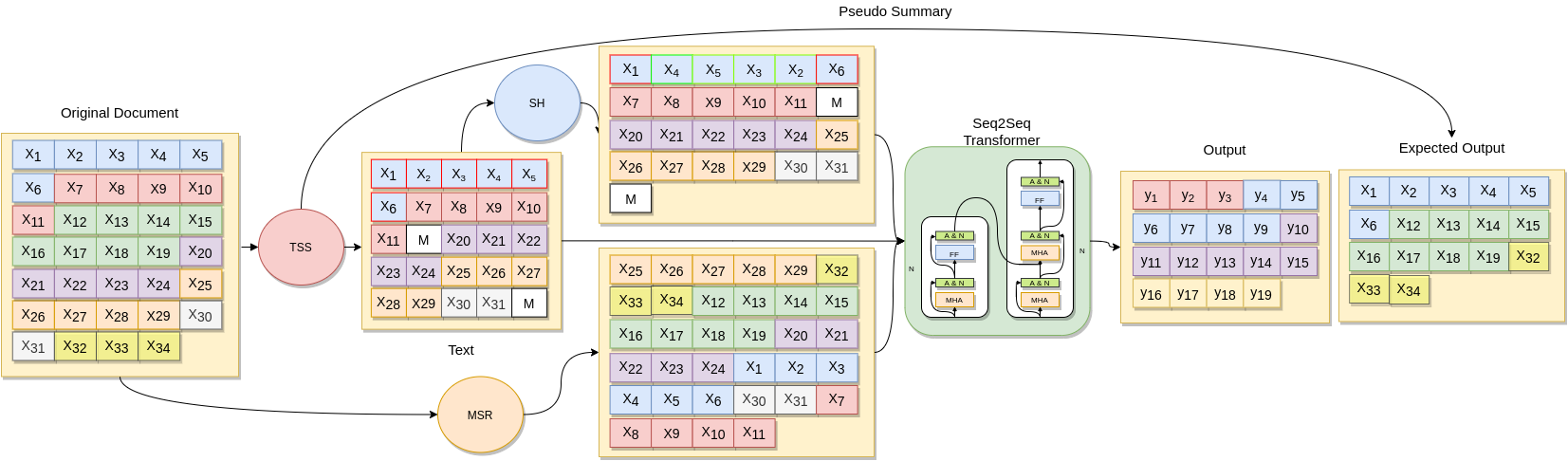}
    \caption{The procedure of making input and output for pre-training Seq2Seq Transformer. TTS selects the salient sentences and divides the original document into text and summary parts. The summary part is the desired output that the Transformer should generate.}
    \label{fig:objectives}
\end{figure*}

\subsection{Pre-training Objectives}
 In this work, we propose new pre-training objectives and compare our models with the closely similar work of PEGASUS \citep{PEGASUS}. We use Transformer encoder-decoder structure and introduce a new combination of masking sentences plus shuffling and reordering objectives. The general procedure of pre-training with the proposed objectives is shown in Figure \ref{fig:objectives}.

\subsubsection*{TSS-ROUGE}
In this objective, we implemented PEGASUS for the Persian language to compare with our proposed models. 
The base architecture of this model is a Transformer encoder-decoder. Instead of masking words, we mask sentences with \textit{<mask>} tokens. In order to generate pseudo summaries as input to this structure, the syntactic-based approach using the ROUGE metric is applied. 

\subsubsection*{TSS-Semantic Similarity (SS)}
This objective takes semantically created pseudo summaries into account. This method is the same as the previous TSS-ROUGE. The semantic-based approach using the modified BERTScore is applied to generate pseudo summaries as input to the structure. The masking criterion is a bit different from TSS-ROUGE. We put only one \textit{<mask>} token for any number of consecutive sentences that should be masked. In this way, the model learns to guess the number of sentences as well. In $20\%$ of the cases, instead of masking a sentence, we keep it in place; this will make the model learn to bring some pieces of the document into the summary. We call the trained model with this objective ARMAN(SS-80).

\subsubsection*{TSS-Shuffling (SH)}
In addition to considering a semantic-based approach for creating a pseudo summary, we apply span shuffling in a sentence and the masking objective together in this objective. In particular, instead of masking sentences $20\%$ of the cases, we shuffle a span of them. The intuition is that the model will learn not to just copy sentences in the final summary and be sensitive to precedence and latency at the span level. We call the trained model with this objective ARMAN(SH).

\subsubsection*{TSS-Modified Sentence Reordering (MSR)}
In this objective, we do masking as the TSS-Semantic Similarity objective does in $90\%$ of documents, and in $10\%$ of other documents, we shuffle all sentences. In the latter, the model should reorder sentences and keep the top $30\%$ of important sentences of the original document according to the semantic scores. The idea behind this method is that the model will learn to arrange the sentences in the correct order in the final summary. Moreover, the model will learn to care about important pieces of the document. In addition to enriching the summary semantically, this work also considers its brevity. We call the trained model with this objective ARMAN(MSR).

\section{Data Collection}
This section introduces the datasets used for pre-training and fine-tuning models and the procedure of cleaning corpora.

\subsection{Pre-training Datasets}
\label{pre-train-datasets}
We merged four large Persian corpora from different sources for pre-training models, which contained formal and informal texts.\\

\noindent
\textbf{irBlogs} \citep{ALEAHMAD2016195} is a collection of 5M+ posts from 600K+ Persian weblogs. Some blogs use informal language for their posts, so this dataset has an enormous amount of informal texts, which could help our models become familiar with this type of Persian speech. \\

\noindent
\textbf{MirasText} \citep{Sabeti2018MirasTextAA} is an automatically produced text corpus for the Persian language by crawling over 250 Persian websites. This corpus contains around 2.8M articles and 1.4B words in all of the articles. \\

\noindent
\textbf{CC100} (\citealp{conneau-etal-2020-unsupervised}; \citealp{wenzek-etal-2020-ccnet}) is a monolingual dataset for 100+ languages constructed from Commoncrawl snapshots. This dataset contains about 111GB of Persian raw text with 13.3B different tokens. 

\noindent
\textbf{YJC News}\footnote{\url{https://github.com/mohammadiahmad/persian-dataset}} is a collection of articles gathered from the Young Journalist Club website\footnote{\url{https://www.yjc.ir/}}. This dataset contains news from various subjects, including 1M+ articles.

\subsection{Downstream Datasets}
For the Summarization task, five datasets were used. All datasets are publicly available and could be used to reproduce our results. Following \citet{grusky2020newsroom}, extractive density and coverage for each summarization dataset has been reported in Appendix \ref{dataset_stats:appendix}. Moreover, we used a Natural Language Understanding (NLU) dataset to test our models' performances on language modeling tasks. \\

\noindent
\textbf{PN-Summary} \citep{farahani2020leveraging} is an abstractive summarization dataset consisting of 93,207 articles of various news categories crawled from six news agency websites. \\

\noindent
\textbf{Wiki-Summary} \citep{Bert2BertWikiSummaryPersian} is a dataset that was extracted from Wikipedia dump files. The main task of this dataset is to generate highlights for each article. There are two versions of this dataset; we used the first version in our experiments, consisting of 56,363 articles and highlight pairs. \\

\noindent
\textbf{VOA Dataset} \citep{Bert2BertFaNewsHeadline} is a medium-sized corpus of 7.9 million words consisting of 38,952 articles of the VOA website\footnote{\url{https://www.voanews.com/}} from 2003 to 2008. The main task that was performed on this dataset was generating a headline for each article. \\

\noindent
\textbf{PerKey} \citep{Doostmohammadi_2018} is a key phrase extraction dataset for the Persian language crawled from six Persian news agencies. There are 553k articles available in this dataset. Some of these articles have summaries, and all of them have titles. \\

\noindent
\textbf{Tebyan Dataset} accumulates 92,289 document-summary pairs that we have collected from the Tebyan website\footnote{\url{https://www.tebyan.net/}}. These articles consist of various subjects and are not limited to news articles. More information about this dataset is provided in Appendix \ref{dataset_stats:appendix}. 

\noindent
\textbf{ParsiNLU} \citep{khashabi2020parsinlu} is a collection of NLU tasks for the Persian language including Textual Entailment, Sentiment Analysis, Question Paraphrasing, Multiple Choice Question Answering, and Reading Comprehension tasks. We have fine-tuned our models on most of them to test their performances on NLU tasks.

\begin{table*}[ht]
\scriptsize
\centering
\renewcommand{\arraystretch}{1.3}
\begin{tabular}{c|cccccc}
\hline
\textbf{Model} & \textbf{PN-Summary} & \textbf{Wiki-Summary} & \textbf{VOA} & \textbf{Perkey(summary)} & \textbf{Perkey(title)} & \textbf{Tebyan} \\
\textbf{} & \textbf{R1/R2/RL} & \textbf{R1/R2/RL} & \textbf{R1/R2/RL} & \textbf{R1/R2/RL} & \textbf{R1/R2/RL} & \textbf{R1/R2/RL} \\
\hline
Transformer\textsubscript{base} & 34.49/16.03/28.91 & 23.96/6.14/17.66 & 31.53/11.71/27.41 & 55.86/43.49/52.22 & 45.33/29.88/42.85 & 23.82/6.79/18.55 \\
PEGASUS\textsubscript{base} & 45.67/27.81/39.71 & 31.98/11.63/23.79 & 47.55/28.68/43.57 & 62.82/51.96/59.48 & 53.99/39.3/51.72 & 37.2/21.23/31.47 \\
ParsBERT\textsubscript{base} & 44.01/25.07/37.76 & 27.34/7.1/\textbf{25.5} & 43.54/24.24/40.76 & - & - & - \\
mT5\textsubscript{small} & 42.25/24.36/35.94 & 15.2/4.73/12.64 & 42.32/25.57/38.99 & 33.88/19.17/28.75 & 28.5/12.55/25.91 & 27.16/12.08/21.27 \\
\hline
ARMAN(SS)\textsubscript{base} & 45.98/28.2/40.09 & 32.27/11.72/23.91 & 47.91/28.9/43.75 & 62.97/52.11/59.64 & 54.18/39.39/51.84 & 37.53/21.73/31.77 \\
ARMAN(SH)\textsubscript{base} & 45.89/28.03/39.89 & 32.04/11.78/23.83 & 46.96/27.88/42.93 & 63.47/52.71/60.16 & 54.5/39.9/52.19 & 37.6/21.77/31.82 \\
ARMAN(MSR)\textsubscript{base} & \textbf{46.19}/\textbf{28.41}/\textbf{40.27} & \textbf{32.48}/\textbf{11.86}/24.08 & \textbf{48.23}/\textbf{29.52}/\textbf{44.27} & \textbf{63.59}/\textbf{52.87}/\textbf{60.3} & \textbf{54.81}/\textbf{40.17}/\textbf{52.51} & \textbf{37.79}/\textbf{21.85}/\textbf{31.98} \\
\hline
\end{tabular}
\caption{\label{results-summarization-rouge} A comparison of results for ARMAN(SS), ARMAN(SH), and ARMAN(MSR) with other pre-trained models on downstream tasks. These results are reported using ROUGE metrics.}
\end{table*}

\begin{table*}[ht]
\scriptsize
\centering
\renewcommand{\arraystretch}{1.3}
\begin{tabular}{c|cccccc}
\hline
\textbf{Model} & \textbf{PN-Summary} & \textbf{Wiki-Summary} & \textbf{VOA} & \textbf{Perkey(summary)} & \textbf{Perkey(title)} & \textbf{Tebyan} \\
\textbf{} & \textbf{P/R/F1} & \textbf{P/R/F1} & \textbf{P/R/F1} & \textbf{P/R/F1} & \textbf{P/R/F1} & \textbf{P/R/F1} \\
\hline
PEGASUS\textsubscript{base} & 79.86/79.67/79.7 & 74.29/71.31/72.64 & 80.84/81.13/80.92 & 86.13/86.01/86.01 & 83.68/83.31/83.45 & 75.26/75.17/75.14 \\
\hline
ARMAN(SS)\textsubscript{base} & 80.08/79.74/79.85 & 74.24/71.48/72.71 & 81.02/81.13/81 & 86.27/86.01/86.09 & 83.65/83.36/83.46 & 75.48/75.32/75.32 \\
ARMAN(SH)\textsubscript{base} & 79.95/79.69/79.76 & 74.25/71.43/72.68 & 80.64/80.91/80.71 & 86.46/86.22/86.29 & 83.85/83.49/83.62 & 75.48/75.28/75.29 \\
ARMAN(MSR)\textsubscript{base} & \textbf{80.14}/\textbf{79.84}/\textbf{79.93} & \textbf{74.67}/\textbf{71.55}/\textbf{72.95} & \textbf{81.1}/\textbf{81.35}/\textbf{81.16} & \textbf{86.54}/\textbf{86.24}/\textbf{86.33} & \textbf{83.93}/\textbf{83.59}/\textbf{83.71} & \textbf{75.49}/\textbf{75.46}/\textbf{75.4} \\
\hline
\end{tabular}
\caption{\label{results-summarization-bertscore}A comparison of results for ARMAN(SS), ARMAN(SH), and ARMAN(MSR) with other pre-trained models on downstream tasks. These results are reported using the original BERTScore metric.}
\end{table*}
\subsection{Preprocessing}
\label{preproccessing_pipeline}
Due to the necessity of a massive amount of data for pre-training of language models, we needed to collect large datasets, but those datasets need to be cleaned. We adopted a heuristic function to produce an automatic pipeline for cleaning our pre-training datasets. First of all, for each document in each dataset, we separated the sentences and removed those which have the following characteristics; 1) sentences with less than five words 2) sentences that do not end with valid Persian end of sentence marks 3) sentences that contain some specific keywords from Persian webpages and javascript codes.

Furthermore, we omitted documents with less than three sentences after the above cleaning. Next, we used the \textit{langdetect}\footnote{\url{https://pypi.org/project/langdetect/}} package to filter out any document which is not in Persian with the probability of 0.99. Lastly, we removed duplicate paragraphs of documents. More information about the size of each corpus after cleaning is reported in Appendix \ref{dataset_stats:appendix}. Our heuristic was inspired by methods from \citet{JMLR:v21:20-074}'s work. This preprocessing procedure only has been used for the pre-training datasets.

\begin{table*}[ht]
\scriptsize
\centering
\renewcommand{\arraystretch}{1.3}
\begin{tabular}{c|cccccc}
\hline
\textbf{Model} & \textbf{PN-Summary} & \textbf{Wiki-Summary} & \textbf{VOA} & \textbf{Perkey(summary)} & \textbf{Perkey(title)} & \textbf{Tebyan} \\
\textbf{} & \textbf{R1/R2/RL} & \textbf{R1/R2/RL} & \textbf{R1/R2/RL} & \textbf{R1/R2/RL} & \textbf{R1/R2/RL} & \textbf{R1/R2/RL} \\
\hline
ARMAN(SS-80)\textsubscript{base} & 45.98/28.2/40.09 & 32.27/11.72/23.91 & \textbf{47.91}/28.9/43.75 & \textbf{62.97}/\textbf{52.11}/\textbf{59.64} & 54.18/39.39/51.84 & 37.53/21.73/31.77 \\
ARMAN(SS-100)\textsubscript{base} & \textbf{46.33}/\textbf{28.57}/\textbf{40.38} & \textbf{32.36}/\textbf{11.78}/\textbf{24.1} & 47.73/\textbf{28.95}/\textbf{43.89} & 62.83/51.92/59.53 & \textbf{54.25}/\textbf{39.51}/\textbf{51.92} & \textbf{37.64}/\textbf{21.78}/\textbf{31.94} \\
\hline
\end{tabular}
\caption{\label{results-SS80-100-summarization-rouge}Comparison of ARMAN(SS-80) and ARMAN(SS-100) results on tasks using ROUGE metrics. }
\end{table*}

\section{Experiments}
In this section, we compare ARMAN with previous
works and conduct several experiments to assess the
performance of the proposed methods. The codes for pre-training and fine-tuning of all models are publicly available on GitHub\footnote{\url{https://github.com/alirezasalemi7/ARMAN}}.

\subsection{Pre-training and Implementation}
Our model is based on Transformer \citep{vaswani_2017} encoder-decoder structure. We pre-trained ARMAN, which contained a 12 layer encoder and a 12 layer decoder with 768 embedding/hidden size, 3072 feed-forward filter size, and 12 self-attention heads. ARMAN and PEGASUS were trained on the mentioned pre-training corpora in section \ref{pre-train-datasets}. The batch size and the training steps of pre-training were set to 128 and 1M, respectively. Adafactor \citep{adafactor} with square root learning rate decay and a dropout rate of 0.1 was used in pre-training and fine-tuning. Pre-training experiments were carried out on the Google Colab platform with TPU v2-8. It took almost 11 days for 1M steps to train ARMAN. Also, we sampled 1M documents from the CC100 dataset and used the SentencePiece Unigram algorithm \citep{kudo} to generate the vocabulary for our models. The size of the vocabulary was 96K in all experiments. 

\subsection{Fine-tuning on Text Summarization}
Abstractive summarization aims to produce a short, fluent, and concise text using advanced natural language techniques to extract essential information from the original document.
We fine-tuned our pre-trained models on six downstream tasks. In all experiments, we set the input length ($L_{input}$) to 512 and output length to 256. Also, we used beam-search as \citet{Wu2016GooglesNM}'s approach with a beam-size of 8 and a length penalty of $0.8$. More information about the experiments' setup is reported in Appendix \ref{hyper_params:appendix}.

Table \ref{results-summarization-rouge} shows results based on standard ROUGE metrics. To compare summaries generated by our models with the state-of-the-art PEGASUS\textsubscript{base} with a text generation evaluation metric, we reported results based on original BERTScore \citep{BERTscore} (using bert-base-multilingual-cased as pre-trained contextual embeddings) in Table \ref{results-summarization-bertscore}. Both tables show the performance improvements of ARMAN(MSR)\textsubscript{base} on all downstream datasets. According to tables \ref{results-summarization-rouge} and \ref{results-summarization-bertscore}, even ARMAN(SS)\textsubscript{base}, our basic proposed method, outperforms PEGASUS\textsubscript{base} in all datasets. These results show that considering the semantic similarity in pre-training objectives is critical in improving the final summary.

In ARMAN(MSR)\textsubscript{base}, we encouraged the model to learn the correct relative orders between sentences by reordering at the sentence level. Results of this model show that the reordering objective gives an improvement in summarization. Our second model, ARMAN(SH)\textsubscript{base}, does not help in improving the quality of summaries. So, we conclude that shuffling at the span level leads to a sub-optimal response, as reported in \citet{JMLR:v21:20-074}.

\subsection{To copy or not to copy!}

\begin{table*}[ht]
\scriptsize
\centering
\renewcommand{\arraystretch}{1.3}
\begin{tabular}{c|cccccc}
\hline
\textbf{Model} & \textbf{PN-Summary} & \textbf{Wiki-Summary} & \textbf{VOA} & \textbf{Perkey(summary)} & \textbf{Perkey(title)} & \textbf{Tebyan} \\
\textbf{} & \textbf{Dens/Cov} & \textbf{Dens/Cov} & \textbf{Dens/Cov} & \textbf{Dens/Cov} & \textbf{Dens/Cov} & \textbf{Dens/Cov} \\
\hline
ARMAN(MSR)\textsubscript{base} & \textbf{8.29486}/\textbf{0.87188} & \textbf{2.55229}/\textbf{0.68437} & 4.59273/0.89648 & \textbf{13.08480}/\textbf{0.84591} & \textbf{2.50826}/\textbf{0.81320} & 18.56819/\textbf{0.86931} \\
PEGASUS\textsubscript{base} & 8.73796/0.87553 & 2.60724/0.68463 & \textbf{4.35264}/\textbf{0.88661} & 13.48538/0.84700 & 2.51945/0.81221 & \textbf{18.23422}/0.87605 \\
\hline
\end{tabular}
\caption{\label{abstractiveness-models}Comparison of ARMAN(MSR) and PEGASUS results on tasks using Density(Dens) and Coverage(Cov) \citep{grusky2020newsroom} metrics. ARMAN has less Density and Coverage in 4 out of 6 datasets.}
\end{table*}

\begin{table*}[!ht]
\scriptsize
\centering
\renewcommand{\arraystretch}{1.3}
\begin{tabular}{c|cccccc}
\hline
\textbf{Model} & \textbf{PN-Summary} & \textbf{Wiki-Summary} & \textbf{VOA} & \textbf{Perkey(summary)} & \textbf{Perkey(title)} & \textbf{Tebyan} \\
\textbf{} & \textbf{F1-T/P-S} & \textbf{F1-T/P-S} & \textbf{F1-T/P-S} & \textbf{F1-T/P-S} & \textbf{F1-T/P-S} & \textbf{F1-T/P-S} \\
\hline
ARMAN(MSR)\textsubscript{base} & \textbf{0.71184}/\textbf{0.82143} & \textbf{0.29325}/\textbf{0.62210} & \textbf{0.58415}/\textbf{0.95609} & 0.58555/0.85361 & \textbf{0.61337}/\textbf{0.90528} & \textbf{0.33835}/\textbf{0.95138} \\
PEGASUS\textsubscript{base} & 0.62983/0.81188 & 0.28139/0.60744 & 0.57440/0.94425 & \textbf{0.64304}/\textbf{0.85691} & 0.54963/0.89825 & 0.30000/0.86349 \\
\hline
\end{tabular}
\caption{\label{factual-models}Comparison of ARMAN(MSR) and PEGASUS results on tasks using F1-Target(F1-T) and Precision-Source(P-S) \citep{nan-etal-2021-entity}. ARMAN has a higher F1-Target and Precision-Source in 5 out of 6 datasets.}
\end{table*}

We observed that PEGASUS\textsubscript{large}\footnote{\url{https://huggingface.co/google/pegasus-large}} tries to copy sentences from the document into a generated summary when it is not fine-tuned on any summarization datasets.The intuition is that when the task is to copy a sentence, and in return for that copying the model gets an extra score, the model becomes biased towards copying the sentences to increase the probability of catching a significant match. In other words, it always copies some sentences from the input to the output with the hope that it will match the output because this yields a decrease in the loss function value.

We set up an experiment to observe the behavior of our models when they are not encouraged to copy sentences of the input into the output. According to semantic score, all proposed methods selected $30\%$ of the top-ranked sentences. In this experiment, we pre-trained ARMAN(SS)\textsubscript{base} with two different values for masking rate in TSS objective; 1) ARMAN(SS-80)\textsubscript{base} masked only $80\%$ of important sentences and left the other $20\%$ unchanged in the input text, 2) ARMAN(SS-100)\textsubscript{base} masked all of the important sentences without copying any sentences from input text into the pseudo summary.

\begin{figure*}[!ht]
    \centering
    \includegraphics[scale=0.38]{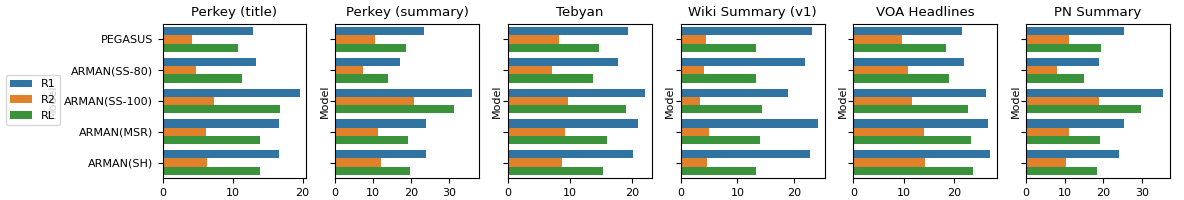}
    \caption{A comparison of results for ARMAN(SS-80), ARMAN(SS-100), ARMAN(SH), ARMAN(MSR), and PEGASUS on zero-shot learning using ROUGE metrics. ARMAN(SS-100) got remarkably better results in most downstream tasks in zero-shot experiments. More details are reported in Appendix \ref{low_resource:appendix}.}
    \label{fig:zeroshot_plot}
\end{figure*}

Results in Figure \ref{fig:zeroshot_plot} show that in a zero-shot setting, ARMAN(SS-100)\textsubscript{base} produces a higher ROUGE score when we do not consider copying in the pre-training objective. Additionally, we fine-tuned ARMAN(SS-100) and ARMAN(SS-80) on downstream tasks. Results in Table \ref{results-SS80-100-summarization-rouge} and Figure \ref{fig:zeroshot_plot} show that ARMAN(SS-100)\textsubscript{base} performs better than ARMAN(SS-80) before and after fine-tuning. Given these results, we used this more effective criteria in our best model, ARMAN(MSR)\textsubscript{base}. 

\begin{figure*}[!ht]
    \centering
    \includegraphics[scale=0.38]{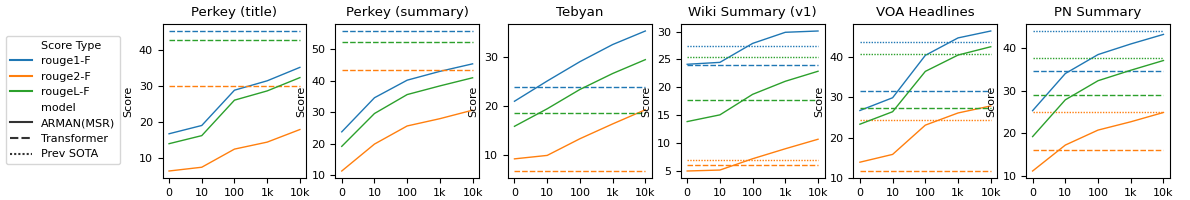}
    \caption{Results of fine-tuning ARMAN(MSR) trained with 0, 10, 100, 1K, 10K examples of each downstream dataset for 2K steps. Also, results of Transformer\textsubscript{base}, which trained on the whole dataset for 150K steps, and previous SOTA (if available) are shown. The results for other models are reported in Appendix \ref{low_resource:appendix}.}
    \label{fig:fewshot_plot}
\end{figure*}

\begin{table*}[ht]
\scriptsize
\centering
\renewcommand{\arraystretch}{1.3}
\begin{tabular}{c|cc|cc|cc|ccc}
\hline
\multicolumn{1}{c}{} & \multicolumn{2}{c}{\textbf{Texual Entailment}} & \multicolumn{2}{c}{\textbf{Question Paraphrasing}} & \multicolumn{2}{c}{\textbf{Sentiment}} & \multicolumn{3}{c}{\textbf{Multiple-Choice Question Answering}} \\
\multicolumn{1}{c}{} & \multicolumn{2}{c}{} & \multicolumn{2}{c}{} & \multicolumn{2}{c}{\textbf{(sentence sent.)}} & \multicolumn{3}{c}{} \\

\hline
 \textbf{Model} & \textbf{natural} & \textbf{translated} & \textbf{natural} & \textbf{translated} & \textbf{food} & \textbf{movie} & \textbf{literature} & \textbf{com-know} & \textbf{math \& logic} \\
 & \textbf{(accuracy)} & \textbf{(accuracy)} & \textbf{(accuracy)} & \textbf{(accuracy)} & \textbf{(F1)} & \textbf{(F1)} & \textbf{(accuracy)} & \textbf{(accuracy)} & \textbf{(accuracy)} \\
\hline
mBERT\textsubscript{base} & 48.7* & 51.6* & 80.4 & 75.3 & 55.2 & 48.6 & 31.1 & 28.6 & 33.8* \\
WikiBERT\textsubscript{base} & 52.8* & 52.6* & 80 & 75.5 & 52 & \textbf{58.5} & 34.0 & \textbf{31.4} & 32.1 \\
ParsBERT\textsubscript{base} & 51.8* & \textbf{53.9}* & 79.4 & 72 & \textbf{59.1} & 56.8 & 35.4 & 29.5 & 32.5* \\
mT5\textsubscript{small} & 51.9	& 51 & 75.2 & 72 & 54.6 & 49.4 & 33.7* & 24.9 & 39.1* \\
PEGASUS\textsubscript{base} & 54.5 & 52.6 & 80 & \textbf{76.1} & 51.9 & 56 & 40 & 27.7 &	45.1 \\
\hline
ARMAN(SS-80)\textsubscript{base} & 54.5 & 50.6 & 82.5 & 74.8 & 51.4 & 47	& 37.7 & 25.7 & 47.7 \\
ARMAN(SS-100)\textsubscript{base} & 54.2 & 53 & 79.9 & 72.8 & 50 & 52.9	& \textbf{41.4} & 27.4 & 43.1 \\
ARMAN(SH)\textsubscript{base} & \textbf{55.5} & 52.9 & \textbf{82.6} & 75.1 & 56.7 & 42 & 34.6 & 28.6 & 45.4\\
ARMAN(MSR)\textsubscript{base} & 54.8 & 51.8 & 79.9 & 75.9 & 52 & 46 & 36.57 & 21.7 & \textbf{49.14} \\
\hline
\end{tabular}
\caption{\label{results-NLU}A comparison of results on ParsiNLU tasks. Some of the reported results (marked with *) in \citet{khashabi2020parsinlu}'s work could not be reproduced according to their policies. So we reported the numbers that we ourselves got using their trained models in our experiments.}
\end{table*}

\subsection{Factual Consistency and Abstractiveness}

From another perspective, we compared the abstractiveness and factual consistency of our best model, ARMAN(MSR), with PEGASUS on downstream summarization tasks because they are important factors for assessing the quality of summaries.

To compare the abstractiveness of models, we calculated the \textit{coverage} and \textit{density} \citep{grusky2020newsroom} of summaries generated by each model. A higher value for \textit{coverage} indicates that the summary uses fewer novel words, and a higher value for \textit{density} is an indicator of a more extractive summary. The average \textit{density} and \textit{coverage} of ARMAN(MSR) and PEGASUS on each dataset are reported in table \ref{abstractiveness-models}. The results show that ARMAN has a lower \textit{density} and \textit{coverage} compared to PEGASUS in 4 out of 6 tasks. Also, in the Tebyan dataset, ARMAN has a higher \textit{density} but lower \textit{coverage}, which means ARMAN uses more novel words compared to PEGASUS. Therefore we conclude that ARMAN’s summaries are more abstractive than PEGASUS.

To compare the factual consistency of models, we calculated \textit{precision-source} and \textit{F1-target} \citep{nan-etal-2021-entity} metrics. While the mentioned metrics evaluate entity-level factual consistency, they still gives considerable information about the factual consistency of models. In order to extract named entities, we used the ParsBERT \citep{farahani2020parsbert} model, which was trained on the PAYMA \citep{Shakery} dataset\footnote{\url{https://huggingface.co/HooshvareLab/bert-fa-base-uncased-ner-peyma}}. The average \textit{precision-source} and \textit{F1-target} of ARMAN(MSR) and PEGASUS on each dataset are reported in Table \ref{factual-models}. The results show that ARMAN has a higher \textit{F1-target} and \textit{precision-source} score than PEGASUS in 5 out of 6 tasks. Therefore, it seems ARMAN is more factually consistent than
PEGASUS. 

\subsection{Zero and Few Shot Summarization}
\label{few-zero-sec}

We studied our models in zero and few shot settings to make abstractive summarization a practical solution for real-world tasks where providing a large supervised collection of training and testing data is laborious. In a zero-shot setting, we pre-trained models on pre-training datasets and examined them on downstream tasks without fine-tuning. Results in Figure \ref{fig:zeroshot_plot} show that our models outperformed PEGASUS. In a few-shot setting, we fed our best model with $10^k$ ($k=1, 2, 3, 4$) examples to study the model's results on low resource scenarios. In this experiment, Transformer\textsubscript{base} and ARMAN(MSR)\textsubscript{base} were trained for 150K and 2K steps, respectively. According to Figure \ref{fig:fewshot_plot}, we observed that in Wiki Summary and VOA datasets, our model has beaten the state-of-the-art model with only seeing 1K samples. In a larger dataset, Perkey, our model did not get a better result than Transformer\textsubscript{base} because it was fine-tuned on the whole dataset with more steps. We conclude that our model gets an acceptable outcome in lower amounts of data and computational resources.

\subsection{NLU Results}
In order to study if ARMAN works well as a language model, we tested our models in Natural Language Understanding (NLU) tasks. According to \citet{khashabi2020parsinlu}, we selected multiple-choice question-answering, textual entailment, sentiment analysis, and question paraphrasing tasks to examine our models' performance on them. For more information about these tasks and datasets, see Appendix \ref{dataset_stats:appendix} and \citet{khashabi2020parsinlu}.


According to the results in Table \ref{results-NLU}, ARMAN(SH)\textsubscript{base} has beaten other models in the natural part of Textual Entailment and Question Paraphrasing. This model learned how to arrange a disordered sentence. Thus, it makes sense why it is powerful in recognizing the same sentences with different written forms. 
In Multiple-Choice QA, our best-performing model achieves the highest accuracy in math and logic questions. Our proposed
model, with semantic similarity and mask-only
approach, surpasses others in literature questions. In the common knowledge task, WikiBERT\textsubscript{base} \citep{pyysalo2020wikibert} outperformed other models because it has been trained over a large Wikipedia dataset. In the Sentiment Analysis task, the proposed models could not achieve acceptable results compared to other models. A more detailed study about the behavior of models on NLU tasks is outside the scope of this work.

\subsection{Human Evaluation}

According to \citet{kryscinski-etal-2019-neural}'s work, we held a human evaluation experiment by considering ROUGE's drawbacks. Our purpose was to determine whether semantic similarity makes better summaries than PEGASUS' GSG in the experiment. Also, we wanted to discover which model is the best from the human's viewpoint.  We selected 30 documents from the PN-Summary dataset and the corresponding generated summaries from PEGASUS, ARMAN(SS-80), and ARMAN(MSR) models. We gave them to 10 participants and asked them to rank the generated summaries from the best to worst similar to \citet{zou-etal-2020-pre}'s work according to fluency, informativeness, and succinctness of the generated summaries. In order to perform statistical tests, we converted rankings into scores ($score= 4-rank$). The experiment result is reported in Table \ref{human-test-results}.  Moreover, we have performed some student t-test between models, and results are reported in Table \ref{human-test-p-val}. Those results show that ARMAN(MSR) is significantly better than other models ($p < 0.05$). Furthermore, results show that ARMAN(SS-80) is not significantly better than PEGASUS but has an extremely small p-value ($0.0507 > 0.05$).

\begin{table}[ht]
\scriptsize
\centering
\renewcommand{\arraystretch}{1.3}
\begin{tabular}{c|cccc}
\hline
\textbf{Model} & \textbf{Rank 1} & \textbf{Rank 2} & \textbf{Rank 3} & \textbf{Score} \\
\hline
PEGASUS & 31\% & 35\% & 34\% & 1.97 \\
ARMAN(SS-80) & 38.33\% & 34.67\% & 27\% & 2.11 \\
ARMAN(MSR) & 50.33\% & 29\% & 20.67\% & 2.29 \\
\hline
\end{tabular}
\caption{\label{human-test-results}Human evaluation results, proportions of model rankings, and average scores. Different models could have the same rankings in tests if they produced the same summary.}
\end{table}

\begin{table}[!h]
\scriptsize
\centering
\renewcommand{\arraystretch}{1.3}
\begin{tabular}{c|ccc}
\hline
\textbf{p-value} & PEGASUS & ARMAN(SS) & ARMAN(MSR) \\
\hline
PEGASUS & - & 0.0507 & $2\times10^{-5}$ \\
ARMAN(SS) & 0.0507 & - & 0.014 \\
ARMAN(MSR) & $2\times10^{-5}$ & 0.014 & - \\
\hline
\end{tabular}
\caption{\label{human-test-p-val}The p-values for models in comparison. ARMAN(MSR) significantly improves results in comparison with ARMAN(SS-80) and PEGASUS ($p < 0.05$).}
\end{table}

\section{Conclusion}
There are few models for generating abstractive summaries in the Persian language. 
This work introduces ARMAN, a Transformer encoder-decoder-based model pre-trained with a new combination of masking sentences with sentence shuffling and reordering objectives. We considered semantic similarities for important sentence selection to make document-summary input data in a self-supervised manner. The results show that the modified sentence selection and reordering model outperforms the most recent SOTA models in all six downstream tasks. Our model achieved a higher score than the previous SOTA with only 1K examples in the case of low supervised sample sizes. Finally, the human evaluation results show significant improvement over the dataset used for this experiment. 

In future work, investigating the effect of using contextual embeddings for selecting salient sentences for producing text and summary pairs might prove necessary. Furthermore, the ability of models on extractive summarization is worth scrutinizing since our objectives select salient sentences, which is similar to extractive summarization.

\section*{Acknowledgements}

We would like to thank the anonymous reviewers for their thoughtful and constructive comments. This research was supported in part by a grant from the Institute for Research in Fundamental Sciences (no. CS 1399-4-286).

\bibliography{anthology,custom}
\bibliographystyle{acl_natbib}


\appendix
\section{Datasets Statistics}
\label{dataset_stats:appendix}

In this section, extra information about downstream datasets and pre-training text corpora is reported. Some of the datasets did not provide any validation split; however, the number of examples in the train/validation/test split and the average length of articles and summaries for each dataset is reported in Table \ref{dataset-statistics:Appendix}. Additionally, the size of pre-training texts corpora before and after preprocessing is reported in Table \ref{pretrain-dataset-statistics:Appendix}.

\begin{figure*}[!h]
    \centering
    \includegraphics[scale=0.35]{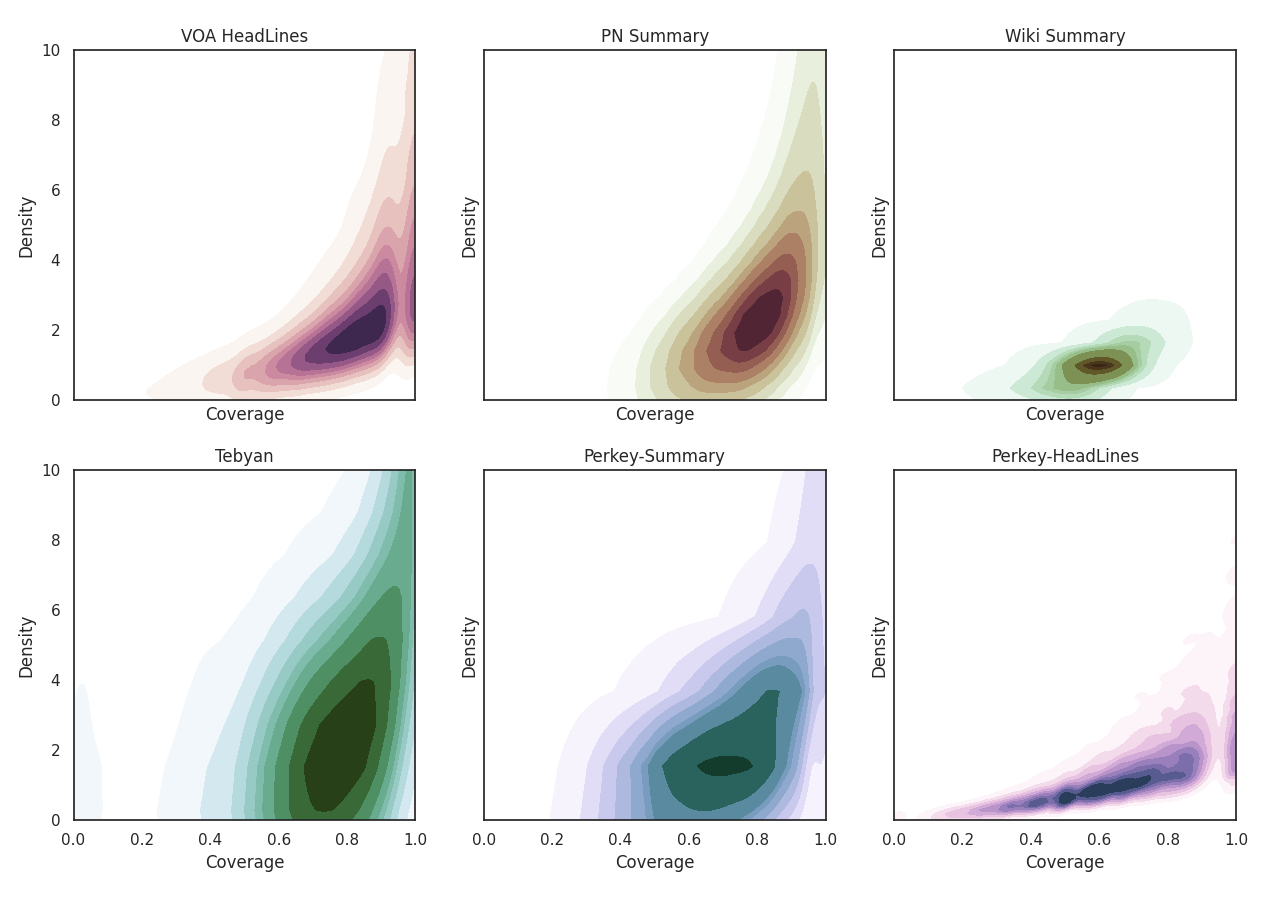}
    \caption{Density and coverage distributions across downstream datasets.}
    \label{fig:density-coverage:Appendix}
\end{figure*}

Following \citet{grusky2020newsroom} and \citet{PEGASUS}, we have plotted the extractive fragment density/coverage plot for each downstream dataset in Figure \ref{fig:density-coverage:Appendix}. \citet{grusky2020newsroom} defined them as  
\begin{gather*}
    \mathrm{COVERAGE(A,S)} = \frac{1}{|S|} \sum_{f \in \mathrm{F(A,S)}}\mathrm{|f|} \\
    \mathrm{DENSITY(A,S)} = \frac{1}{|S|} \sum_{f \in \mathrm{F(A,S)}}\mathrm{|f|^2}
\end{gather*}
where $A$ is an article and $S$ is the corresponding summary, and $F(A,S)$ is the set of shared sequences of tokens in $A$ and $S$. The density for extractive summaries is higher than more abstractive summaries. Lower coverage shows the novelty of text fragments in summary. Figure \ref{fig:density-coverage:Appendix} shows that our downstream datasets range from more extractive summaries to more abstractive ones.

Tebyan dataset contains articles and summaries from a well-known Persian lifestyle website that includes various articles from different categories. In order to produce the Tebyan dataset, we have crawled 100K pages of their site. We removed all HTML tags using \textit{beautifulsoup4}\footnote{\url{https://pypi.org/project/beautifulsoup4/}} for each page, and each page's primary content was stored with the author's provided summary, and paragraphs were separated with a newline character. Lastly, we have used Langdetect\footnote{\url{https://pypi.org/project/langdetect/}} to remove articles that were not in Persian. After to this procedure, 92,289 articles and summaries were collected, so we separated them into three parts, $85\%$ for train, $7.5\%$ for validation, and $7.5\%$ for test split.

We have tested ARMAN on NLU tasks with the ParsiNLU \citep{khashabi2020parsinlu}, which is a Persian NLU dataset. This dataset consists of 5 main tasks and translation as an extra task. In Table \ref{parsinlu-stats:Appendix}, the number of examples for train/validation/test of ParsiNLU is reported. It should be noted that we did not test ARMAN on the Reading Comprehension of this dataset due to resource leakage. The Sentiment Analysis task of this dataset has two sub-tasks; sentence-level sentiment and aspect-based sentiment of a sentence. We have tested ARMAN on the sentence-level sentiment task.  The Sentiment Analysis task of this dataset has two subtasks; sentence-level sentiment and aspect-based sentiment of a sentence. We have tested ARMAN on the sentence-level sentiment task. For Question Paraphrasing and Textual Entailment, this dataset contains two subtasks; sentences written by humans and sentences translated from English datasets into Persian, so we have reported the accuracy of models for each subtask separately.

\section{ARMAN Hyper Parameters and Training Settings}
\label{hyper_params:appendix}

In this section, we have described pre-training and fine-tuning parameters and settings. In Table \ref{pretrain-setting:Appendix}, pre-training settings for ARMAN\textsubscript{base} are reported. Tables \ref{finetuning-sum-setting:Appendix} and \ref{finetuning-transformer-sum-setting:Appendix} contain information about settings used in fine-tuning ARMAN\textsubscript{base} and Transformer\textsubscript{base} on summarization tasks.  Also, the fine-tuning settings for NLU tasks are reported in Table \ref{finetuning-nlu-setting:Appendix}. Finally, we have reported each model's parameter counts used in summarization tasks in Table \ref{model-total-params:Appendix}.

\section{Low Resource Numbers and Settings}
\label{low_resource:appendix}

Table \ref{finetuning-low-resource-setting:Appendix} contains information about fine-tuning settings for low resource experiments in Section \ref{few-zero-sec}. The numbers in Figures \ref{fig:zeroshot_plot} and \ref{fig:fewshot_plot} are reported in Table \ref{ARMAN-MSR-low-resource:Appendix}. We did not report the results of low resource experiments for ARMAN(SH), ARAMN(SS-80), ARMAN(SS-100), and PEGASUS in the main part of the paper, but they are reported in Tables \ref{ARMAN-SH-low-resource:Appendix}, \ref{ARMAN-SS-80-low-resource:Appendix}, \ref{ARMAN-SS-100-low-resource:Appendix}, and \ref{PEGASUS-low-resource:Appendix}. 

\begin{table*}[!h]
\centering
\scriptsize
\renewcommand{\arraystretch}{1.3}
\begin{tabular}{ccc}
\hline
\textbf{Pre-train Corpus/Dataset} & \textbf{Original Corpus/Dataset Size} & \textbf{Cleaned Corpus/Dataset Size} \\
\hline
irBlogs & 7.1GB & 2.6GB \\
MirasText & 15.7GB & 6.8GB \\
YJC News & 3GB & 2.3GB \\
CC100 & 111GB & 53GB \\
\hline
Total & 136.8GB & 64.7GB \\
\hline
\end{tabular}
\caption{\label{pretrain-dataset-statistics:Appendix}
Size of pre-training text corpora in GB for each corpus before and after cleaning.
}
\end{table*}

\begin{table*}[!h]
\centering
\scriptsize
\renewcommand{\arraystretch}{1.3}
\begin{tabular}{cccc}
\hline
\textbf{Task} & \textbf{Number of Train Examples} & \textbf{Number of  Validation Examples} & \textbf{Number of Test Examples} \\
\hline
Reading Comprehension & 600 & 125 & 575 \\
Multiple-Choice & 1271 & 139 & 1050 \\
Sentiment Analysis & 1894 & 235 & 294 \\
Textual Entailment & 756 & 271 & 1751 \\
Question Paraphrasing & 1830 & 898 & 1916 \\
\hline
\end{tabular}
\caption{\label{parsinlu-stats:Appendix} Task name and the number of examples for ParsiNLU dataset.
}
\end{table*}

\begin{table*}[!h]
\centering
\scriptsize
\renewcommand{\arraystretch}{1.3}
\begin{tabular}{cccccc}
\hline
\textbf{Dataset} & \textbf{Train Count} & \textbf{Validation Count} & \textbf{Test Count} & \textbf{Article Average Length} & \textbf{Summary Average Length}\\
\hline
PN-Summary & 82022 & 5592 & 5593 & 335 & 31 \\
Wiki-Summary & 45653 & 5073 & 5637 & 425 & 82 \\
VOA & 31550 & 3506 & 3896 & 179 & 11 \\
Perkey (summary) & 42077 & - & 19796 & 218 & 28 \\
Perkey (title) & 526445 & - & 24930 & 224 & 11 \\
Tebyan & 78445 & 6922 & 6922 & 819 & 37 \\
\hline
\end{tabular}
\caption{\label{dataset-statistics:Appendix}
The number of articles and summaries and the average length of them for each downstream dataset (lengths are reported in words count).
}
\end{table*}

\begin{table*}[!hb]
\centering
\scriptsize
\renewcommand{\arraystretch}{1.3}
\begin{tabular}{c|ccccccc}
\hline
\textbf{Model} & \textbf{Learning rate} & \textbf{Label Smoothing} & \textbf{Steps} & \textbf{Batch Size} & \textbf{Objective} & \textbf{Max Input Length} & \textbf{Max Output Length} \\
\hline
PEGASUS\textsubscript{base} & 0.01 & 0.0 & 1M & 128 & Ind-Orig & 512 & 128 \\
\hline
ARMAN(SS)\textsubscript{base} & 0.01 & 0.0 & 1M & 128 & TSS & 512 & 128 \\
ARMAN(SH)\textsubscript{base} & 0.01 & 0.0 & 1M & 128 & TSS+Shuffling & 512 & 128 \\
ARMAN(MSR)\textsubscript{base} & 0.01 & 0.0 & 1M & 128 & TTS+MSR & 512 & 128 \\
\hline
\end{tabular}
\caption{\label{pretrain-setting:Appendix} Pre-training settings for ARMAN\textsubscript{base} models. We have used PEGASUS\textsubscript{large} \citep{PEGASUS} settings for maximum input and output length since they have searched for the best setting.}
\end{table*}

\begin{table*}[!hb]
\centering
\scriptsize
\renewcommand{\arraystretch}{1.3}
\begin{tabular}{c|cccccccc}
\hline
\textbf{Dataset} & \textbf{Learning rate} & \textbf{Label Smoothing} & \textbf{Steps} & \textbf{Batch Size} & \textbf{Beam Size} & \textbf{Beam alpha} & \textbf{Max Input} & \textbf{Max Output} \\
\hline
Perkey (summary) & $5\times 10^{-4}$ & 0.1 & 50K & 128 & 8 & 0.8 & 512 & 256 \\
Perkey (title) & $5\times 10^{-4}$ & 0.1 & 50K & 128 & 8 & 0.8 & 512 & 256 \\
PN-Summary & $5\times 10^{-4}$ & 0.1 & 50K & 128 & 8 & 0.8 & 512 & 256 \\
Tebyan & $5\times 10^{-4}$ & 0.1 & 50K & 128 & 8 & 0.8 & 512 & 256 \\
VOA & $5\times 10^{-4}$ & 0.1 & 20K & 64 & 8 & 0.8 & 512 & 256 \\
Wiki-Summary (v1) & $5\times 10^{-4}$ & 0.1 & 50K & 64 & 8 & 0.8 & 512 & 256 \\
\hline
\end{tabular}
\caption{\label{finetuning-sum-setting:Appendix} Fine-tuning settings for ARMAN\textsubscript{base} models on downstream summarization tasks and datasets.}
\end{table*}

\begin{table*}[!hb]
\centering
\scriptsize
\renewcommand{\arraystretch}{1.3}
\begin{tabular}{c|cccccccc}
\hline
\textbf{Dataset} & \textbf{Learning rate} & \textbf{Label Smoothing} & \textbf{Steps} & \textbf{Batch Size} & \textbf{Beam Size} & \textbf{Beam alpha} & \textbf{Max Input} & \textbf{Max Output} \\
\hline
Perkey (summary) & $5\times 10^{-4}$ & 0.1 & 150K & 128 & 8 & 0.8 & 512 & 256 \\
Perkey (title) & $5\times 10^{-4}$ & 0.1 & 150K & 128 & 8 & 0.8 & 512 & 256 \\
PN-Summary & $5\times 10^{-4}$ & 0.1 & 150K & 128 & 8 & 0.8 & 512 & 256 \\
Tebyan & $5\times 10^{-4}$ & 0.1 & 150K & 128 & 8 & 0.8 & 512 & 256 \\
VOA & $5\times 10^{-4}$ & 0.1 & 150K & 64 & 8 & 0.8 & 512 & 256 \\
Wiki-Summary (v1) & $5\times 10^{-4}$ & 0.1 & 150K & 64 & 8 & 0.8 & 512 & 256 \\
\hline
\end{tabular}
\caption{\label{finetuning-transformer-sum-setting:Appendix} Fine-tuning settings for Transformer\textsubscript{base} models on downstream summarization tasks and datasets.}
\end{table*}

\begin{table*}[!hb]
\centering
\scriptsize
\renewcommand{\arraystretch}{1.3}
\begin{tabular}{c|cccccccc}
\hline
\textbf{Dataset} & \textbf{Learning rate} & \textbf{Label Smoothing} & \textbf{Steps} & \textbf{Batch Size} & \textbf{Beam Size} & \textbf{Beam alpha} & \textbf{Max Input} & \textbf{Max Output} \\
\hline
Multiple-Choice & $5\times 10^{-4}$ & 0.1 & 20K & 48 & 8 & 0.8 & 512 & 256 \\
Sentiment Analysis & $5\times 10^{-4}$ & 0.1 & 20K & 48 & 8 & 0.8 & 512 & 256 \\
Textual Entailment & $5\times 10^{-4}$ & 0.1 & 20K & 48 & 8 & 0.8 & 512 & 256 \\
Question Paraphrasing & $5\times 10^{-4}$ & 0.1 & 20K & 48 & 8 & 0.8 & 512 & 256 \\
\hline
\end{tabular}
\caption{\label{finetuning-nlu-setting:Appendix} Fine-tuning settings for ARMAN\textsubscript{base} models on NLU tasks. Batch size  48 was chosen to be the same as other models that were trained on those tasks. We have converted the classification problem into the text to text problems.}
\end{table*}

\begin{table*}[!hb]
\centering
\scriptsize
\renewcommand{\arraystretch}{1.3}
\begin{tabular}{ccc}
\hline
\textbf{Model} & \textbf{Parameters} & \textbf{Transformer Type} \\
\hline
ARMAN\textsubscript{base} & 223M & \citet{vaswani_2017}'s Encoder-Decoder\\
Transformer\textsubscript{base} & 223M & \citet{vaswani_2017}'s Encoder-Decoder \\
ParsBERT\textsubscript{base} \citep{farahani2020leveraging} & 221M & \citet{10.1162/tacl_a_00313}'s Encoder-Decoder \\
PEGASUS\textsubscript{base} \citep{PEGASUS} & 223M & \citet{vaswani_2017}'s Encoder-Decoder \\
mT5\textsubscript{small} \citep{Xue2020mT5AM} & 300M & \citet{vaswani_2017}'s Encoder-Decoder \\
\hline
\end{tabular}
\caption{\label{model-total-params:Appendix}
Parameters count for each tested model that was used for summarization. Reported numbers are in millions.}
\end{table*}

\begin{table*}[!h]
\centering
\scriptsize
\renewcommand{\arraystretch}{1.3}
\begin{tabular}{c|cccccccc}
\hline
\textbf{Dataset} & \textbf{Learning rate} & \textbf{Label Smoothing} & \textbf{Steps} & \textbf{Batch Size} & \textbf{Beam Size} & \textbf{Beam alpha} & \textbf{Max Input} & \textbf{Max Output} \\
\hline
Perkey (summary) & $5\times 10^{-4}$ & 0.1 & 2K & 128 & 8 & 0.8 & 512 & 256 \\
Perkey (title) & $5\times 10^{-4}$ & 0.1 & 2K & 128 & 8 & 0.8 & 512 & 256 \\
PN-Summary & $5\times 10^{-4}$ & 0.1 & 2K & 128 & 8 & 0.8 & 512 & 256 \\
Tebyan & $5\times 10^{-4}$ & 0.1 & 2K & 128 & 8 & 0.8 & 512 & 256 \\
VOA & $5\times 10^{-4}$ & 0.1 & 2K & 64 & 8 & 0.8 & 512 & 256 \\
Wiki-Summary (v1) & $5\times 10^{-4}$ & 0.1 & 2K & 64 & 8 & 0.8 & 512 & 256 \\
\hline
\end{tabular}
\caption{\label{finetuning-low-resource-setting:Appendix} Fine-tuning settings for ARMAN\textsubscript{base} and PEGASUS\textsubscript{base} models on downstream summarization tasks and datasets for low resource experiments.}
\end{table*}

\begin{table*}[!h]
\scriptsize
\centering
\renewcommand{\arraystretch}{1.3}
\begin{tabular}{c|cccccc}
\hline
\textbf{examples} & \textbf{PN-Summary} & \textbf{Wiki-Summary} & \textbf{VOA} & \textbf{Perkey(summary)} & \textbf{Perkey(title)} & \textbf{Tebyan} \\
\textbf{} & \textbf{R1/R2/RL} & \textbf{R1/R2/RL} & \textbf{R1/R2/RL} & \textbf{R1/R2/RL} & \textbf{R1/R2/RL} & \textbf{R1/R2/RL} \\
\hline
0 & 25.28/11.09/19.18 & 24.14/5.05/13.89 & 26.69/13.89/23.3 & 23.83/11.39/19.22 & 16.65/6.26/13.88 & 20.93/9.25/15.85 \\
10 & 34.01/17.19/27.83 & 24.48/5.22/15.09 & 29.85/15.82/26.39 & 34.66/19.94/29.65 & 18.96/7.31/16.13 & 25.03/9.95/19.4 \\
100 & 38.47/20.71/32.32 & 27.87/7.24/18.76 & 40.35/23.07/36.38 & 40.24/25.7/35.67 & 28.81/12.37/26.03 & 28.95/13.33/23.32 \\
1K & 40.96/22.66/34.78 & 29.86/9.03/21.1 & 44.67/26.08/40.42 & 43.04/28.01/38.42 & 31.43/14.32/28.61 & 32.42/16.33/26.55 \\
10K & 43.21/24.85/37.07 & 30.09/10.73/22.89 & 46.38/27.82/42.48 & 45.43/30.71/41 & 35.18/17.83/32.32 & 35.17/19.2/29.36 \\
\hline
\end{tabular}
\caption{\label{ARMAN-MSR-low-resource:Appendix} Low resource results of ARMAN(MSR) from Figures \ref{fig:zeroshot_plot} and \ref{fig:fewshot_plot}. By less than 1000 examples, ARMAN(MSR) has beaten the previous SOTA on VOA and Wiki-Summary datasets. Also, 10K examples and 2K  fine-tuning steps got comparable results with previous SOTA in the Pn-Summary dataset.}
\end{table*}

\begin{table*}[!h]
\scriptsize
\centering
\renewcommand{\arraystretch}{1.3}
\begin{tabular}{c|cccccc}
\hline
\textbf{examples} & \textbf{PN-Summary} & \textbf{Wiki-Summary} & \textbf{VOA} & \textbf{Perkey(summary)} & \textbf{Perkey(title)} & \textbf{Tebyan} \\
\textbf{} & \textbf{R1/R2/RL} & \textbf{R1/R2/RL} & \textbf{R1/R2/RL} & \textbf{R1/R2/RL} & \textbf{R1/R2/RL} & \textbf{R1/R2/RL} \\
\hline
0 & 24.08/10.34/18.37 & 22.73/4.58/13.24 & 27.13/14.14/23.67 & 23.91/12.07/19.6 & 16.69/6.36/13.9 & 20.1/8.74/15.23 \\
10 & 34.71/17.63/28.51 & 24.6/5.43/15.27 & 33.88/18.5/30.06 & 38.27/23.88/33.61 & 22.93/8.97/20.15 & 25.91/11.4/20.27 \\
100 & 38.67/20.67/32.49 & 27.41/7.42/19.03 & 41.05/23.39/37.03 & 41.34/26.31/36.58 & 26.83/11.28/24.09 & 29.13/13.49/23.52 \\
1K & 40.95/22.78/34.7 & 30/8.68/20.75 & 44.22/25.11/39.77 & 43.11/28.06/38.45 & 31.2/14.17/28.28 & 33.1/17.24/27.31 \\
10K & 43.07/24.84/37.05 & 29.83/10.43/22.58 & 46.8/27.87/42.86 & 45.19/30.43/40.76 & 34.79/17.53/31.83 & 34.71/18.83/28.97 \\
\hline
\end{tabular}
\caption{\label{ARMAN-SH-low-resource:Appendix} Low resource results of ARMAN(SH). By less than 1000 examples, ARMAN(SH) has beaten the previous SOTA on VOA and Wiki-Summary datasets. Also, 10K examples and 2K  fine-tuning steps got comparable results with previous SOTA in the Pn-Summary dataset.}
\end{table*}

\begin{table*}[!h]
\scriptsize
\centering
\renewcommand{\arraystretch}{1.3}
\begin{tabular}{c|cccccc}
\hline
\textbf{examples} & \textbf{PN-Summary} & \textbf{Wiki-Summary} & \textbf{VOA} & \textbf{Perkey(summary)} & \textbf{Perkey(title)} & \textbf{Tebyan} \\
\textbf{} & \textbf{R1/R2/RL} & \textbf{R1/R2/RL} & \textbf{R1/R2/RL} & \textbf{R1/R2/RL} & \textbf{R1/R2/RL} & \textbf{R1/R2/RL} \\
\hline
0 & 18.92/7.96/15.04 & 21.87/4.14/13.22 & 21.8/10.81/19.02 & 17.19/7.36/14.02 & 13.33/4.8/11.31 & 17.67/7.05/13.7 \\
10 & 37.1/19.18/30.54 & 24.84/5.99/16.45 & 33.84/17.6/30.09 & 35.36/21.14/30.58 & 25.17/10.47/22.21 & 27.74/12.97/22.39 \\
100 & 39.26/21.2/33.21 & 27.54/7.28/18.89 & 41.17/23.15/37.31 & 40.6/25.89/36.22 & 28.54/12.23/25.72 & 30.83/15.42/25.33 \\
1K & 40.51/22.38/34.43 & 29.75/8.66/20.72 & 44.09/25.51/39.9 & 42.64/27.58/38.05 & 30.73/13.87/27.89 & 32.27/16.31/26.49 \\
10K & 43.03/24.82/36.91 & 29.36/10.2/22.33 & 46.88/27.96/42.91 & 44.94/30.18/40.51 & 34.53/17.31/31.65 & 34.78/18.74/28.94 \\
\hline
\end{tabular}
\caption{\label{ARMAN-SS-80-low-resource:Appendix} Low resource results of ARMAN(SS-80). By less than 1000 examples, ARMAN(SS-80) has beaten the previous SOTA on VOA and Wiki-Summary datasets. Also, 10K examples and 2K  fine-tuning steps got comparable results with previous SOTA in the Pn-Summary dataset.}
\end{table*}

\begin{table*}[!h]
\scriptsize
\centering
\renewcommand{\arraystretch}{1.3}
\begin{tabular}{c|cccccc}
\hline
\textbf{examples} & \textbf{PN-Summary} & \textbf{Wiki-Summary} & \textbf{VOA} & \textbf{Perkey(summary)} & \textbf{Perkey(title)} & \textbf{Tebyan} \\
\textbf{} & \textbf{R1/R2/RL} & \textbf{R1/R2/RL} & \textbf{R1/R2/RL} & \textbf{R1/R2/RL} & \textbf{R1/R2/RL} & \textbf{R1/R2/RL} \\
\hline
0 & 35.53/18.91/29.78 & 18.86/3.42/14.4 & 26.18/11.51/22.75 & 36.15/20.89/31.37 & 19.58/7.34/16.73 & 22.04/9.67/19.04 \\
10 & 38.49/20.79/32.49 & 24.05/6.47/17.59 & 33.45/16.42/29.66 & 38.55/23.05/33.72 & 24.85/10.18/21.74 & 28.53/14.13/23.84 \\
100 & 39.26/21.19/33.17 & 27.76/7.46/19.31 & 41.52/22.97/37.52 & 40.71/25.27/35.84 & 28.63/12.27/25.7 & 30.32/14.29/24.73 \\
1K & 41.25/22.89/35.16 & 30.15/8.88/21.01 & 44.88/25.58/40.67 & 42.97/27.75/38.36 & 31.38/14.32/28.52 & 32.85/17.02/27.26 \\
10K & 43.51/25.28/37.42 & 29.48/10.16/22.31 & 46.98/28.33/43.07 & 45/30.08/40.52 & 35.29/17.95/32.42 & 34.95/19/29.22 \\
\hline
\end{tabular}
\caption{\label{ARMAN-SS-100-low-resource:Appendix} Low resource results of ARMAN(SS-100). By less than 1000 examples, ARMAN(SS-100) has beaten the previous SOTA on VOA and Wiki-Summary datasets. Also, 10K examples and 2K  fine-tuning steps got comparable results with previous SOTA in the Pn-Summary dataset.}
\end{table*}

\begin{table*}[!t]
\scriptsize
\centering
\renewcommand{\arraystretch}{1.3}
\begin{tabular}{c|cccccc}
\hline
\textbf{examples} & \textbf{PN-Summary} & \textbf{Wiki-Summary} & \textbf{VOA} & \textbf{Perkey(summary)} & \textbf{Perkey(title)} & \textbf{Tebyan} \\
\textbf{} & \textbf{R1/R2/RL} & \textbf{R1/R2/RL} & \textbf{R1/R2/RL} & \textbf{R1/R2/RL} & \textbf{R1/R2/RL} & \textbf{R1/R2/RL} \\
\hline
0 & 25.32/11.25/19.45 & 23.11/4.49/13.24 & 21.51/9.56/18.36 & 23.34/10.41/18.68 & 12.93/4.2/10.81 & 19.27/8.16/14.64 \\
10 & 35.18/17.65/28.99 & 24.06/5.49/16.36 & 30.14/14.83/26.5 & 34.49/19.2/29.36 & 16.75/6.14/14.29 & 26.81/11.39/20.98 \\
100 & 37.94/20/31.71 & 27.27/6.81/18.32 & 40.23/22.68/36.46 & 39.97/25.06/35.4 & 26.47/11.08/23.75 & 28.86/13.12/23.27 \\
1K & 39.91/21.88/33.82 & 29.46/8.61/20.61 & 42.67/23.8/38.53 & 42.11/27.09/37.48 & 29.97/13.46/27.19 & 31.87/16.01/26.2 \\
10K & 42.27/24.06/36.2 & 29.3/10.13/22.22 & 46.04/27.15/41.91 & 44.72/29.97/40.37 & 33.98/16.93/31.11 & 34.59/18.73/28.91 \\
\hline
\end{tabular}
\caption{\label{PEGASUS-low-resource:Appendix} Low resource results of PEGASUS. }
\end{table*}

\section{Samples}
\begin{figure*}[!b]
    \centering
    \includegraphics[scale=0.5]{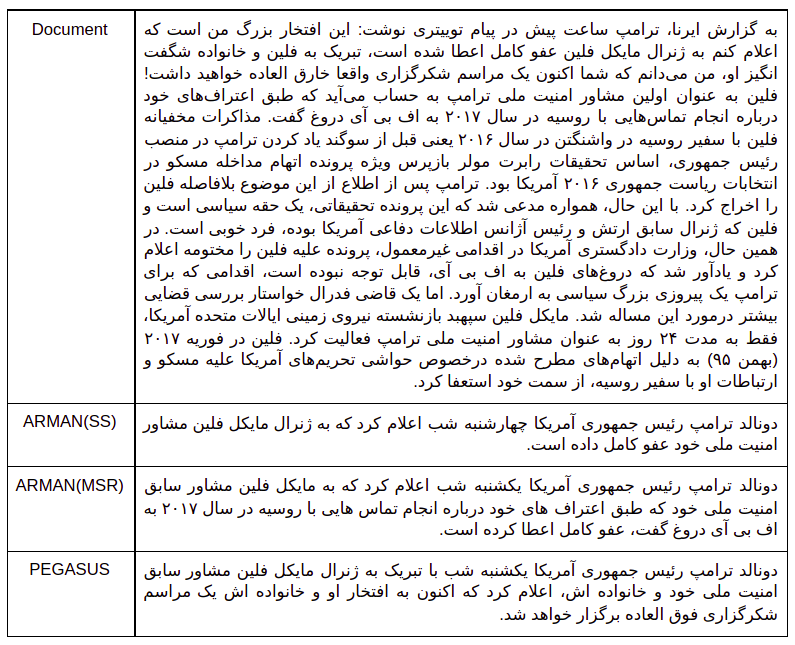}
    \caption{The First sample of models' generated summaries in human evaluation tests.}
    \label{fig:gen_sum_sample1:Appendix}
\end{figure*}

\begin{figure*}[!b]
    \centering
    \includegraphics[scale=0.5]{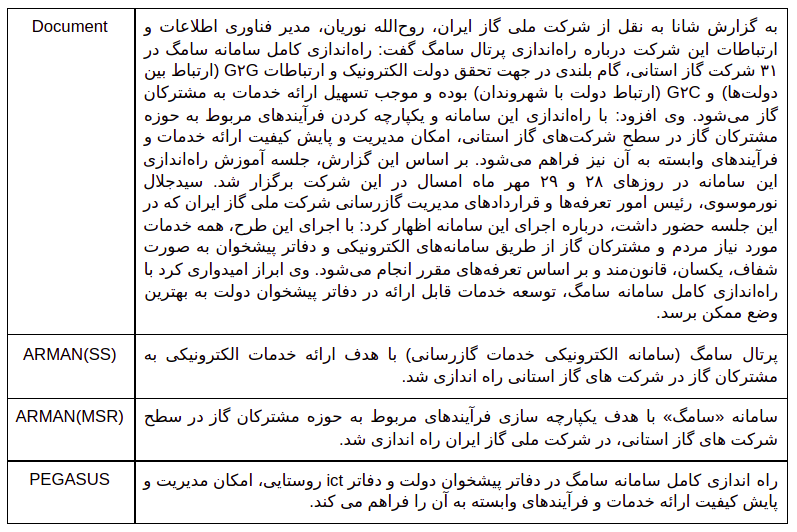}
    \caption{The Second sample of models' generated summaries in human evaluation tests.}
    \label{fig:gen_sum_sample2:Appendix}
\end{figure*}

Two samples of ARMAN(SS), ARMAN(MSR), and PEGASUS generated summaries that were used in the human evaluation test are shown in Figures \ref{fig:gen_sum_sample1:Appendix} and \ref{fig:gen_sum_sample2:Appendix}. More than 50\% of participants believed that ARMAN(MSR)'s summaries were the best among all models in the human evaluation test, which shows that its summaries have high quality.

\end{document}